\documentclass[letterpaper]{article} 
\usepackage{aaai2026}  
\usepackage{times}  
\usepackage{helvet}  
\usepackage{courier}  
\usepackage[hyphens]{url}  
\usepackage{graphicx} 
\usepackage{natbib}  
\usepackage{caption} 
\usepackage{algorithm}
\usepackage{algorithmic}
\usepackage{amsmath}
\usepackage{amsfonts}

\usepackage[table]{xcolor}
\usepackage{multirow}
\usepackage{caption}
\usepackage{subcaption}
\usepackage{tabularx}
\newcolumntype{C}{>{\centering\arraybackslash}X}

\usepackage{booktabs}

\usepackage{newfloat}
\usepackage{listings}
\DeclareCaptionStyle{ruled}{labelfont=normalfont,labelsep=colon,strut=off} 
\floatstyle{ruled}
\newfloat{listing}{tb}{lst}{}
\floatname{listing}{Listing}
\title{Revisiting Cross-Architecture Distillation: Adaptive Dual-Teacher Transfer for Lightweight Video Models}
\author{
    Ying Peng\textsuperscript{\rm 1},
    Hongsen Ye\textsuperscript{\rm 1},
    Changxin Huang\textsuperscript{\rm 2},
    Xiping Hu\textsuperscript{\rm 3},
    Jian Chen\textsuperscript{1,}\footnotemark[1],
    Runhao Zeng\textsuperscript{3,}\footnotemark[1]  
}

\affiliations{
    \textsuperscript{\rm 1}South China University of Technology,
    \textsuperscript{\rm 2}Shenzhen University, \\
    \textsuperscript{\rm 3}Artificial Intelligence Research Institute, Shenzhen MSU-BIT University

    \{py.yingpeng, runhaozeng.cs\}@gmail.com, huxp@bit.edu.cn, ellachen@scut.edu.cn
}

\usepackage{bibentry}

\begin{document}

\maketitle
\renewcommand{\thefootnote}{\fnsymbol{footnote}}
\footnotetext[1]{Corresponding Author}
\renewcommand{\thefootnote}{\arabic{footnote}}

\begin{abstract}
Vision Transformers (ViTs) have achieved strong performance in video action recognition, but their high computational cost limits their practicality. Lightweight CNNs are more efficient but suffer from accuracy gaps. Cross-Architecture Knowledge Distillation (CAKD) addresses this by transferring knowledge from ViTs to CNNs, yet existing methods often struggle with architectural mismatch and overlook the value of stronger homogeneous CNN teachers. To tackle these challenges, we propose a Dual-Teacher Knowledge Distillation framework that leverages both a heterogeneous ViT teacher and a homogeneous CNN teacher to collaboratively guide a lightweight CNN student. We introduce two key components: (1) Discrepancy-Aware Teacher Weighting, which dynamically fuses the predictions from ViT and CNN teachers by assigning adaptive weights based on teacher confidence and prediction discrepancy with the student, enabling more informative and effective supervision; and (2) a Structure Discrepancy-Aware Distillation strategy, where the student learns the residual features between ViT and CNN teachers via a lightweight auxiliary branch, focusing on transferable architectural differences without mimicking all of ViT’s high-dimensional patterns. Extensive experiments on benchmarks including HMDB51, EPIC-KITCHENS-100, and Kinetics-400, demonstrate that our method consistently outperforms state-of-the-art distillation approaches, achieving notable performance improvements with a maximum accuracy gain of 5.95\% on HMDB51.
\end{abstract}


\section{Introduction}
Video action recognition has received increasing attention in recent years due to its wide applications in intelligent surveillance, human–computer interaction, and autonomous driving. Vision Transformers (ViTs)~\cite{arnab2021vivit,liu2022video,tong2022videomae}, with their powerful global modeling capability, have achieved remarkable performance in this domain and have become one of the mainstream high-accuracy model families. However, the large parameter size and high computational cost of ViTs hinder their deployment on resource-constrained edge devices. In contrast, lightweight convolutional neural networks (CNNs) ~\cite{feichtenhofer2020x3d,kondratyuk2021movinets}, are more suitable for practical deployment due to their efficient inference and hardware-friendliness, but they suffer from notable performance gaps compared to ViTs. Bridging this gap while maintaining model efficiency remains a core challenge.

To address this issue, Cross-Architecture Knowledge Distillation (CAKD) has emerged as a promising solution, which aims to transfer knowledge from a powerful ViT teacher to a lightweight CNN student. However, due to the fundamental differences between ViTs and CNNs in terms of inductive biases, receptive fields, and architectural designs, their intermediate representations are difficult to align semantically and spatially. To overcome this challenge, existing CAKD methods often introduce feature projection modules to map heterogeneous features into a shared latent space. For example, CAKD~\cite{liu2022cross} incorporates complex projection modules involving Partially Cross Attention and Group-wise Linear Transformation for feature alignment, while OFA-KD~\cite{hao2023one} introduces auxiliary exits in the student model to project intermediate features into the logits space, thus bypassing structural mismatches. Although these approaches alleviate feature-level misalignment to some extent, they heavily rely on intricate alignment modules, suffer from limited training stability and scalability, and more importantly, overlook a valuable source of knowledge—homogeneous CNN teachers, which are structurally similar to the student.

Although the overall accuracy of a homogeneous CNN teacher may be lower than that of a ViT, our preliminary experiments reveal that incorporating a CNN teacher can sometimes yield even greater performance gains. We attribute this to the structural consistency between the CNN teacher and student, which facilitates more effective feature transfer and guidance. This observation motivates us to rethink the ViT-dominant paradigm in CAKD and explore whether a more collaborative and compatible distillation framework can be designed to fully leverage both heterogeneous and homogeneous teachers.

To this end, we propose a novel Dual-Teacher Knowledge Distillation Framework, where a ViT teacher and a CNN teacher collaboratively guide a lightweight CNN student. However, introducing dual teachers raises two key challenges: \textbf{1) how can the student effectively select and balance the supervision from two structurally dissimilar teachers during training?} and \textbf{2) how can we extract complementary and architecture-relevant knowledge from both teachers, rather than forcing direct alignment with heterogeneous features?}

\textbf{For the first challenge}, we perform a statistical analysis\footnote{Please refer to the Appendix for more details.} and observe that distillation gains are most significant when the teacher shows high confidence (i.e., low entropy) and there is a large discrepancy between teacher and student predictions (e.g., low cosine similarity). These conditions indicate that the teacher offers reliable yet novel knowledge for the student. Based on this, we propose a \textbf{Discrepancy-Aware Teacher Weighting (DATW)} mechanism, which computes the distillation target as a weighted combination of the two teachers’ logits. The weights are dynamically assigned per sample, considering both teacher confidence and prediction discrepancy, allowing adaptive and informative supervision from the more helpful teacher.

\textbf{To tackle the second challenge}, we argue that directly aligning student features with those of a ViT teacher is suboptimal due to substantial architectural and representational differences. Instead, we ask: can the difference between heterogeneous teachers offer more targeted supervision? Since both the ViT and CNN teachers are more expressive than the student, the residual between their features naturally captures the student’s structural deficiency—especially the global context strength of ViT. We propose a \textbf{Structure Discrepancy-Aware Distillation} mechanism that guides the student to predict this residual. A lightweight non-local module~\cite{Wang_2018_CVPR} takes the student’s features as input and learns to approximate the difference between ViT and CNN teacher features. This encourages the student to internalize the architectural gap and extract the ViT’s complementary knowledge. The auxiliary prediction branch is only used during training and is detached at inference.

We validate the effectiveness of our method on HMDB51, EPIC-KITCHENS-100, and Kinetics-400.  Results demonstrate that the student models trained with our framework not only outperform existing methods, but even exceed the performance of both ViT and CNN teachers.

Our main contributions are summarized as follows:
\begin{itemize}
    \item We propose a dual-teacher cross-architecture distillation framework for video action recognition, which is the first to jointly leverage both a heterogeneous ViT teacher and a homogeneous CNN teacher, breaking the ViT-only paradigm in existing CAKD methods.
    \item We introduce a discrepancy-aware teacher weighting mechanism that adaptively balances supervision from both teachers, enabling sample-specific guidance based on confidence and prediction disagreement.
    \item We design a structure discrepancy-aware distillation method that encourages the student to learn transferable differences between ViT and CNN features, improving global representation capability without incurring inference overhead. Extensive experiments demonstrate that our method significantly outperforms existing state-of-the-art distillation techniques.
\end{itemize}

\begin{figure}[t]          
  \centering             
  \includegraphics[width = \columnwidth]{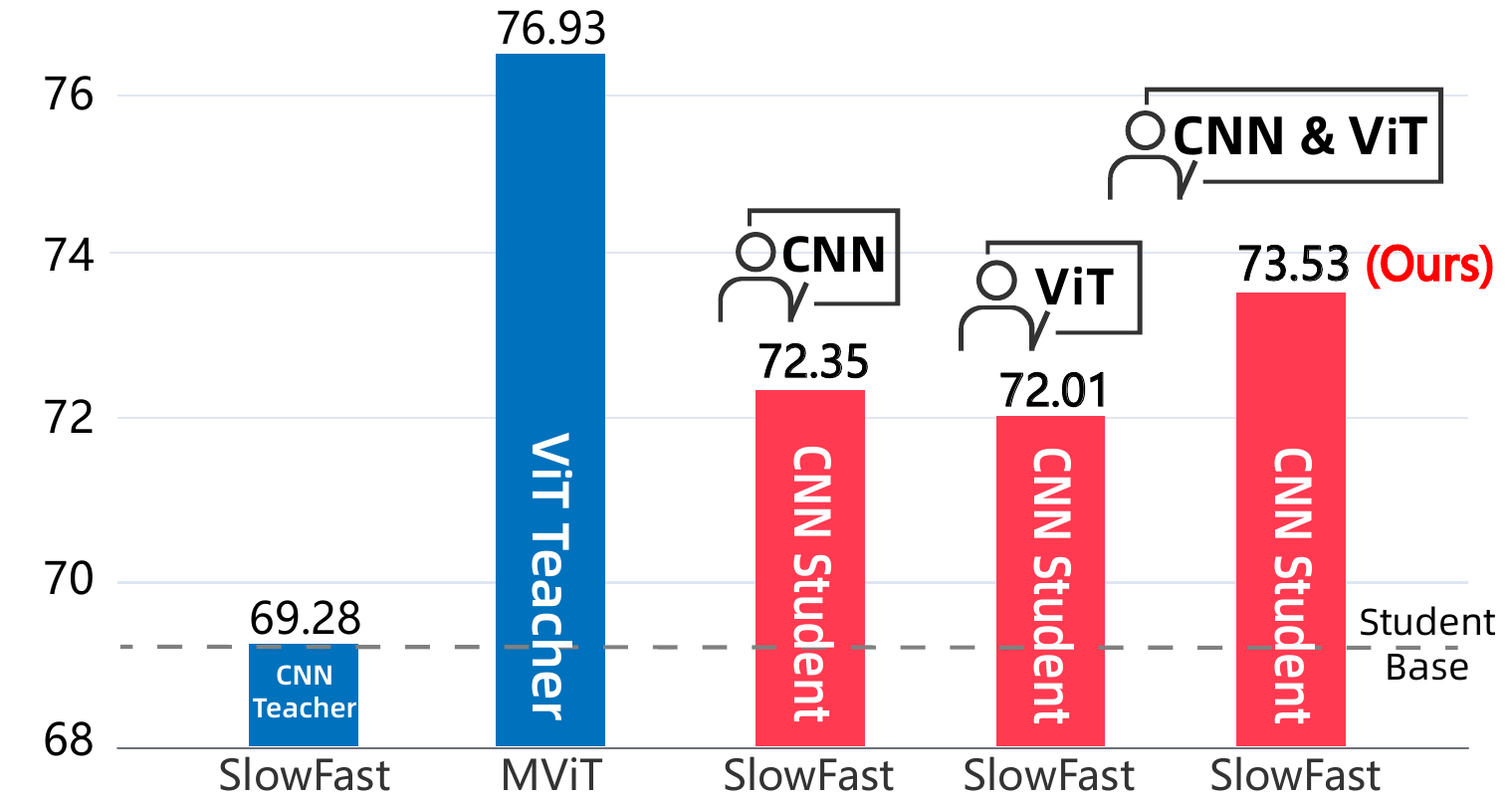} 
  \vspace{-15pt}
\caption{Motivation for our dual-teacher framework. A preliminary study shows that distillation from a weaker but structurally aligned CNN teacher can sometimes lead to better student performance than from a stronger, heterogeneous ViT teacher. This finding motivates the integration of both teacher types and highlights the critical role of architectural compatibility in cross-architecture knowledge transfer.}
  \label{fig:fig1} 
\end{figure}

\section{Related Work}
\label{sec:related_work}
\subsection{Video Action Recognition}
Video action recognition aims to classify human actions in videos. Early works like I3D~\cite{Carreira_2017_CVPR}, R(2+1)D~\cite{Tran_2018_CVPR}, and SlowFast~\cite{Feichtenhofer_2019_ICCV} advanced spatio-temporal modeling via 3D convolutions and dual-pathway designs. To further reduce overhead, methods such as TSM~\cite{lin2019tsm}, X3D~\cite{feichtenhofer2020x3d}, and MoViNet~\cite{kondratyuk2021movinets} propose lightweight temporal modules and compact architectures optimized for mobile devices.

In parallel, Vision Transformers (ViTs) have shown strong performance in video tasks by modeling long-range dependencies through spatio-temporal self-attention~\cite{bertasius2021space, Liu_2022_CVPR, li2022mvitv2}, often surpassing CNNs. Hybrid models like UniFormer~\cite{li2022uniformer} further integrate local convolution and global Transformer reasoning for enhanced performance.

However, the high computational and memory demands of ViTs hinder their scalability in real-world applications. In contrast, CNNs are more deployment-friendly due to their low latency, efficient memory usage, and hardware support (e.g., CUDA, TensorRT), yet they typically lag behind ViTs in modeling global context, resulting in inferior performance on challenging benchmarks. This has sparked growing interest in using ViTs as teachers to improve lightweight CNNs through knowledge distillation. However, cross-architecture distillation for video action recognition remains underexplored, particularly in settings involving both structurally aligned CNNs and dissimilar ViTs. This motivates our study on how to effectively combine the complementary strengths of both teacher types to guide efficient CNN students.

\begin{figure*}[t]
  \centering
  \includegraphics[width=\textwidth]{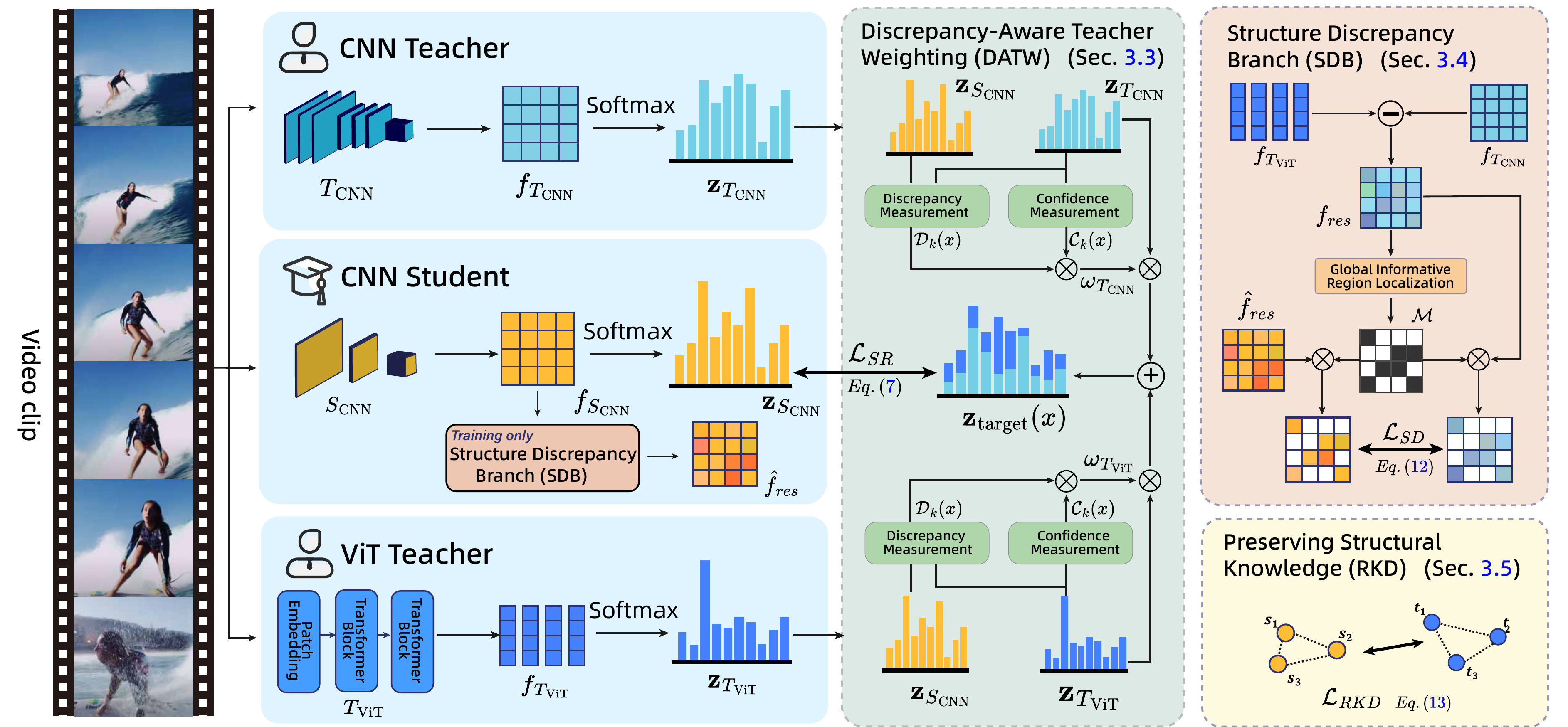}
  \vspace{-15pt}
  \caption{An overview of our dual-teacher distillation framework.
Given a lightweight CNN student $S_{\text{CNN}}$, a ViT teacher $T_\text{ViT}$, and a CNN teacher $T_\text{CNN}$, we distill complementary knowledge via three components: 1) \textbf{Discrepancy-Aware Teacher Weighting} mechanism measures teacher confidence $\mathcal{C}_k(x)$ and prediction discrepancy $\mathcal{D}_k(x)$ for each sample, which are integrated to generate adaptive weights $\omega_k(x)$ for combining teacher logits. This enables the student to prioritize informative and reliable supervision on a per-sample basis. 2) \textbf{Structure Discrepancy Branch} predicts the feature residual $f_{T_\text{ViT}} - f_{T_\text{CNN}}$ using a Non-local module, enabling the student to capture ViT-specific global context cues. 
At inference, only $S_{\text{CNN}}$ is used, introducing no extra computational overhead. 3) \textbf{Relational Knowledge Distillation} transfers architecture-agnostic structural knowledge. 
} 
  \label{fig:fig2} 
\end{figure*}

\subsection{Knowledge Distillation (KD)}
KD ~\cite{hinton2015distilling,wang2023masked,yu2024select,li2025adaptive,liu2025noisy} has been extensively studied to transfer knowledge from large models to compact ones. Existing approaches span logits-based~\cite{li2023curriculum,mirzadeh2020improved,zhang2018deep,sun2024logit}, feature-based~\cite{chen2021distilling,chen2022knowledge,tian2019contrastive,wang2024generative}, and relation-based~\cite{park2019relational, yim2017gift,peng2019correlation} strategies. However, most are designed for homogeneous architectures (e.g., CNN-to-CNN) and assume strong structural alignment between teacher and student.

Cross-architecture distillation (CAKD), especially from ViTs to CNNs, is more challenging due to the architectural gap. Recent CAKD works~\cite{liu2022cross, hao2023one,zhao2023cross} attempt to bridge this via projection modules or shared latent spaces. Yet, these designs often require complex feature alignment and ignore the representational strengths of structurally aligned CNN teachers.

Notably, in video action recognition, cross-architecture KD remains underexplored. Existing methods primarily distill from a single ViT teacher and often overlook the value of combining CNN teachers for structure-compatible guidance. Our approach addresses these gaps by introducing a dual-teacher framework that integrates both ViT and CNN teachers. We propose structure-aware mechanisms to adaptively fuse global and local inductive biases, allowing lightweight CNN students to benefit from heterogeneous supervision.

\section{Proposed Method}
\subsection{Preliminaries and Motivation}
Cross-Architecture Knowledge Distillation (CAKD) typically aims to transfer knowledge from a powerful ViT teacher ($T_{\text{ViT}}$) to a lightweight CNN student ($S_{\text{CNN}}$). Given an input video $V \in \mathbb{R}^{T \times H \times W \times C}$, where $T, H, W,$ and $C$ denote the number of frames, height, width, and channels of the video clip, respectively. The ViT teacher first embeds the input into a sequence of patch tokens and processes them through Transformer blocks, yielding features $f_{T_{\text{ViT}}} \in \mathbb{R}^{N \times D}$, where $N$ is the number of tokens and $D$ is the embedding dimension. In contrast, the CNN student extracts spatial-temporal features $f_{S_{\text{CNN}}} \in \mathbb{R}^{T' \times H' \times W' \times D'}$ via 3D convolutions. Both features are then forwarded to task heads to produce final logit outputs $\mathbf{z}_{T_{\text{ViT}}}$ and $\mathbf{z}_{S_{\text{CNN}}}$.

CAKD typically operates on two levels: \textbf{logits-based} distillation and \textbf{feature-based} distillation. The former encourages the student to mimic the softened output distribution of the teacher to learn inter-class relationships:
\begin{equation}
    \mathcal{L}_{\text{KD}}^{\text{logits}}(\mathbf{z}_{T_{\text{ViT}}}, \mathbf{z}_{S_{\text{CNN}}}) = \tau^2 \cdot \text{KL}(\sigma(\mathbf{z}_{T_{\text{ViT}}} / \tau) \parallel \sigma(\mathbf{z}_{S_{\text{CNN}}} / \tau)),
    \label{eq:kd_logits}
\end{equation}
where $\sigma(\cdot)$ is the softmax function and $\tau$ is the temperature parameter. The latter aims to encourage the student to mimic the teacher’s intermediate representations. The loss is typically defined as:
\begin{equation}
    \mathcal{L}_{\text{KD}}^{\text{feat}}(f_{T_{\text{ViT}}}, f_{S_{\text{CNN}}}) = \| \phi(f_{S_{\text{CNN}}}) - f_{T_{\text{ViT}}} \|^2
    \label{eq:kd_feat}
\end{equation}
where $\phi(\cdot)$ is an optional projection module used to align feature dimensions when necessary.

\subsubsection{Motivation}
Existing CAKD methods face two key limitations. \textbf{First}, overlooking structurally homogeneous CNN teachers, which, despite lower accuracy than ViTs, can more effectively guide CNN students due to shared architectural priors (see Figure~\ref{fig:fig1}). \textbf{Second}, enforcing full imitation of ViT features, risking architectural noise and suboptimal learning. These insights motivate us to integrate both teacher types, enabling the student to benefit from the structural alignment of CNNs and the representational strength of ViTs through complementary, targeted supervision.

\subsection{General Scheme}

To address the aforementioned limitations, we propose a \textbf{Dual-Teacher Knowledge Distillation (DT-KD)} framework that jointly leverages a heterogeneous ViT teacher ($T_{\text{ViT}}$) and a homogeneous CNN teacher ($T_{\text{CNN}}$) to guide a lightweight CNN student ($S_{\text{CNN}}$). This framework addresses two key challenges:
\textbf{C1}--how to adaptively balance supervision from structurally dissimilar teachers, and
\textbf{C2}--how to extract complementary, architecture-relevant knowledge without enforcing direct feature alignment.

To tackle \textbf{C1}, we introduce \textbf{Discrepancy-Aware Teacher Weighting (DATW)}, which assigns sample-wise weights based on teacher confidence and student-teacher prediction discrepancy. This enables dynamic teacher cooperation at the logits level, forming the loss $\mathcal{L}_{\text{SR}}$.

To address \textbf{C2}, we propose \textbf{Structure Discrepancy-Aware Distillation (SDD)}. Instead of forcing the student to mimic ViT features directly, we introduce a lightweight auxiliary branch $g_{\text{SDB}}(\cdot)$ during training to predict the residual between the ViT and CNN teacher features—capturing architecture-specific knowledge.

Finally, a \textbf{Relational Knowledge Distillation (RKD)} loss $\mathcal{L}_{\text{RKD}}$ further preserves high-level relational structure from the ViT teacher. These components work jointly to transfer knowledge adaptively and effectively.

\subsection{Discrepancy-Aware Teacher Weighting}

A central challenge in dual-teacher distillation lies in determining how a student model should effectively balance and select supervision from two structurally dissimilar teachers To improve the logits-level distillation, we aim to replace the fixed teacher target $\mathbf{z}_{T_{\text{ViT}}}$ in Eq. (\ref{eq:kd_logits}) with a sample-adaptive combination of both teacher logits:
\begin{equation}
\mathbf{z}_{\text{target}}(x) = \omega_{\text{CNN}}(x) \cdot \mathbf{z}_{T_{\text{CNN}}}(x) + \omega_{\text{ViT}}(x) \cdot \mathbf{z}_{T_{\text{ViT}}}(x),
\end{equation}
where $\omega_{\text{CNN}}(x)$ and $\omega_{\text{ViT}}(x)$ are adaptive weights for each input $x$.
Through an in-depth analysis of our preliminary experiments, we find that the most beneficial knowledge for distillation typically arises when the teacher’s predictions exhibit both \textbf{1)} high confidence and \textbf{2)} large discrepancy from the student’s predictions (Detailed analysis is provided in the Appendix). Based on this insight, we propose the \textbf{Discrepancy-Aware Teacher Weighting (DATW)} mechanism, which quantifies teacher quality using two metrics and translates them into soft arbitration weights in a differentiable manner. 

\subsubsection{Measuring Teacher Confidence} We measure the confidence of teacher $k \in \{\text{CNN, ViT}\}$ on sample $x$ using the normalized negative entropy of its softmax prediction:
\begin{equation}
\mathcal{C}_k(x) = \frac{1}{\log N} \sum_{i=1}^{N} p_{k,i}(x) \log p_{k,i}(x),
\end{equation}
where $p_k(x) = \text{Softmax}(\mathbf{z}_k(x) / \tau)$ is the temperature-scaled prediction, and $N$ is the number of classes. Lower entropy implies higher certainty in the teacher’s prediction.

\subsubsection{Measuring Student–Teacher Discrepancy}
To assess the potential informational gain from each teacher, we compute the cosine distance between student's and teacher's logits:
\begin{equation}
\mathcal{D}_k(x) = 1 - \frac{\mathbf{z}_{S_{\text{CNN}}}(x) \cdot \mathbf{z}_k(x)}{\|\mathbf{z}_{S_{\text{CNN}}}(x)\| \|\mathbf{z}_k(x)\|}.
\end{equation}
A larger value of $\mathcal{D}_k(x)$ implies greater disagreement and hence a higher chance for the teacher to provide complementary knowledge.

\subsubsection{Combining Confidence and Discrepancy}
We define a guidance efficacy score $s_k(x)$ by multiplying the above two metrics and normalize the scores across both teachers to obtain the final arbitration weights:
\begin{align}
    \omega_k(x) &= \frac{s_k(x)}{\sum_{j \in \{\text{CNN, ViT}\}} s_j(x) + \epsilon}, \label{eq:weights} \\
    s_k(x) &= \mathcal{C}_k(x) \cdot \mathcal{D}_k(x), \notag
\end{align}
where $\epsilon$ is a small constant for numerical stability. Using the weighted teacher logits $\mathbf{z}_{\text{target}}(x)$, the final logits-level distillation loss becomes:
\begin{equation}
\mathcal{L}_{\text{SR}} = \tau^2 \cdot \mathbb{E}_{x} \left[
\text{KL} \left( \sigma(\mathbf{z}_{\text{target}}(x)/\tau) \,\|\, \sigma(\mathbf{z}_{S_{\text{CNN}}}(x)/\tau) \right)
\right].
\end{equation}

\subsection{Structure Discrepancy-Aware Distillation}

Transferring feature-level knowledge from a ViT teacher to a CNN student is challenging due to their divergent architectural priors~\cite{liu2022cross,hao2023one}. ViTs excel at capturing global dependencies through self-attention, while CNNs mainly extract local patterns via convolution. Forcing the student to mimic ViT features holistically may result in poor alignment and limited performance gains due to mismatched representational formats.

\subsubsection{Learning via Feature Residuals} To mitigate this, we propose a Structure Discrepancy-Aware Distillation (SDD) mechanism, which reframes the learning target as the feature residual between a ViT and a structurally aligned CNN teacher. This residual emphasizes global cues encoded by the ViT but missing from the CNN, offering a more focused and complementary supervision signal for the student. Formally, the residual is defined as:
\begin{equation}
    f_{\text{res}} = \phi_{\text{ViT}}(f_{T_{\text{ViT}}}) - \phi_{\text{CNN}}(f_{T_{\text{CNN}}}),
    \label{eq:residual_feature}
\end{equation}
where $\phi(\cdot)$ is an optional projection function for aligning feature dimensions. This design leverages the CNN teacher as a reference anchor, allowing the student to learn the ViT's distinct structural cues without being overwhelmed by direct, incompatible supervision.

To localize supervision to the most informative regions, we propose a Dual-Masked Sparse Distillation strategy, combining two spatial masks:
First, the residual-based mask $\mathcal{M}_{\text{res}}$ identifies regions where the ViT and CNN teachers differ significantly, by thresholding the residual map $f_{\text{res}}$ at its $\zeta$-quantile:
\begin{equation}
    \mathcal{M}_{\text{res}}(i, j) =
    \begin{cases}
        1, & \text{if } f_{\text{res}}(i, j) > Q_{\zeta} \\
        0, & \text{otherwise}
    \end{cases}.
    \label{eq:mask_res}
\end{equation}
Second, to suppress irrelevant background noise and enhance semantic concentration, we define an activation-based mask $\mathcal{M}_{\text{act}}$ that highlights semantically relevant regions based on high average activations from the CNN teacher:
\begin{equation}
\mathcal{M}_{\text{act}}(i, j) =
\begin{cases}
1, & \text{if } \frac{1}{C} \sum_{c=1}^{C} f_{T_{\text{CNN}}}^{(c)}(i, j) > Q_{\rho} \\
0, & \text{otherwise}
\end{cases}
\label{eq:mask_act}
\end{equation}
where $f_{T_{\text{CNN}}}^{(c)}(i, j)$ denotes the activation value at spatial location $(i, j)$ in channel $c$, and $C$ is the total number of channels. We apply supervision only over the intersection $\mathcal{M}_{\text{res}} \cap \mathcal{M}_{\text{act}}$, focusing learning on structurally distinctive and semantically salient regions.

\begin{table*}[t]
\centering
\renewcommand{\arraystretch}{1.3}
\caption{State-of-the-art comparison on HMDB51. Our dual-teacher method consistently outperforms existing distillation methods. Notably, our distilled X3D-S student achieves \textbf{77.06\%} accuracy, surpassing its powerful MViTv2-S teacher (76.93\%) while achieving a \textbf{96\%} reduction in FLOPs and an \textbf{89\%} reduction in parameters (3.76M vs. 34.5M).}
\vspace{-7pt}
\label{tab:hmdb51}
\scalebox{0.8}{
\setlength{\tabcolsep}{15pt} 
\begin{tabular}{c@{\hspace{12pt}}r@{\hspace{10pt}}|cccc|rc}
\hline 
\multicolumn{2}{c|}{Method} &\multicolumn{4}{c|}{Only ViT Teacher} &\multicolumn{2}{c}{ViT Teacher+CNN Teacher}\\

ViT Teacher  & Student w/o distillstion   & ATKD & MGD & DKD  & PixelKD & CNN Teacher  & \cellcolor{gray!20}\textbf{Ours} \\
\hline

\multirow{3}{*}{\shortstack{MViTv2-S\\(76.93)}} &  
MoViNet--A2 (71.76)   & 72.88 & 72.48 & 75.16  & 70.20 &MoViNet-A2 (71.76)& \cellcolor{gray!20}\textbf{76.41} \ \\
 & SlowFast-R50 (69.28)  & 68.17 & 67.84 & 69.28  & 72.03 &SlowFast-R50 (69.28)& \cellcolor{gray!20}\textbf{75.23}\\
 & X3D-S (72.42) & 64.64 & 70.72 & 73.66  & 75.16 &X3D-M (69.80)& \cellcolor{gray!20}\textbf{77.06} \\
\hline
\end{tabular}
}
\end{table*}

\subsubsection{Auxiliary Learning with Structure Discrepancy Branch (SDB)}
To avoid entangling this learning with the student’s core prediction path, we introduce a lightweight auxiliary module called the Structure Discrepancy Branch (SDB) $g_{\text{SDB}}$. This branch operates during training only and is detached at inference. It takes the student’s backbone features $f_{S_{\text{CNN}}}$ as input and predicts the ViT-CNN residual:
\begin{equation}
\hat{f}_{\text{res}} = g_{\text{SDB}}(f_{S_{\text{CNN}}}) = \mathrm{SE} \big( \mathrm{NL}(f_{S_{\text{CNN}}}) \big).
\end{equation}
Here, \(\mathrm{NL}(\cdot)\) is a non-local block that models global spatial dependencies, mimicking ViT-style attention, while \(\mathrm{SE}(\cdot)\) is a squeeze-and-excitation module that recalibrates channel-wise responses to enhance feature expressiveness. Isolating these components in the SDB allows the student to learn architecture-specific global information without disrupting its primary learning pathway.

The final distillation loss is computed as the masked MSE between the predicted and ground-truth residuals:
\begin{equation}
    \mathcal{L}_{SD} =  \left\| \left( g_{\text{SDB}}(f_{S_{\text{CNN}}}) - f_{\text{res}} \right) \odot \mathcal{M}_{\text{res}} \odot \mathcal{M}_{\text{act}} \right\|_2^2.
    \label{eq:sd_loss}
\end{equation}
By modeling ViT–CNN structural differences explicitly and restricting supervision to the most informative regions, SDB offers a principled, efficient pathway for the student to selectively absorb global inductive knowledge that would otherwise be difficult to transfer across architectures.

\subsection{Preserving Structural Knowledge with RKD}
As previously mentioned, directly aligning intermediate features between ViT and CNN models is often suboptimal due to their inherently different architectural priors and feature semantics. To alleviate this, we incorporate Relational Knowledge Distillation (RKD) \cite{park2019relational} to transfer architecture-agnostic structural knowledge.

RKD bypasses absolute feature matching and instead focuses on preserving the pairwise relational structure among samples in the feature space. This approach captures higher-order inter-sample dependencies, which are more transferable across heterogeneous architectures.

Specifically, for each sample pair $(i,j)$ in a mini-batch, we compute their Euclidean distances in both teacher and student feature spaces, denoted as $d_T(i,j)$ and $d_S(i,j)$, and normalize by the batch-wise mean distances $\bar{d}_T$, $\bar{d}_S$. The RKD loss is then defined as:
\begin{equation}
\mathcal{L}_{\text{RKD}} = \frac{1}{|\mathcal{P}|} \sum_{(i,j) \in \mathcal{P}} 
\text{SmoothL1} \left( 
\frac{d_S(i,j)}{\bar{d}_S}, 
\frac{d_T(i,j)}{\bar{d}_T} 
\right)
\end{equation}
where $\mathcal{P}$ is the set of all distinct sample pairs in the batch. This encourages the student to preserve the geometric structure induced by the teacher, without requiring explicit feature alignment.

\subsection{Training and Inference Details}
\subsubsection{Training Objective} The student model is trained end-to-end by minimizing a comprehensive objective function that integrates the standard supervised loss with our proposed multi-level distillation losses. The final objective, $\mathcal{L}_{\text{Total}}$, is formulated as:
\begin{equation}
    \mathcal{L}_{\text{Total}} = \mathcal{L}_{\text{\text{CE}}} + \alpha \mathcal{L}_{\text{SR}} + \beta \mathcal{L}_{\text{SD}} + \gamma \mathcal{L}_{\text{RKD}}
    \label{eq:final_objective}
\end{equation}
where $\mathcal{L}_{\text{CE}}$ is the cross-entropy loss on the ground-truth labels, while $\alpha$, $\beta$, and $\gamma$ are hyperparameters that balance the contributions of our proposed distillation components.

\subsubsection{Inference Efficiency}
During inference, the student model operates independently without requiring access to either teacher models or auxiliary modules. In particular, the auxiliary structure discrepancy branch (SDB), which is introduced only during training to guide feature learning via $\mathcal{L}_{SD}$, is entirely discarded after training. As a result, the student retains its original architecture, with no additional parameters or computational overhead introduced at test time.

\begin{table*}[tbp]
\centering
\renewcommand{\arraystretch}{1.3}
\caption{State-of-the-art comparison on EPIC-Kitchens. We adopt SlowFast-R50 as the CNN Teacher model for comparisons.}
\vspace{-7pt}
\label{tab:epic}
\scalebox{0.8}{
\setlength{\tabcolsep}{15pt} 
\begin{tabular}{c@{\hspace{12pt}}c@{\hspace{10pt}}c@{\hspace{10pt}}c|cccc|cc}
\hline
 
\multicolumn{4}{c|}{Method} &

\multicolumn{4}{c|}{Only ViT Teacher} &
\multicolumn{2}{c}{ViT Teacher+CNN Teacher} \\

Task & Student Model  & ViT Teacher  & Student
               & ATKD  & MGD
               & DKD 
               & PixelKD & CNN Teacher & \cellcolor{gray!20}\textbf{Ours} \\

\hline

Verb &\multirow{2}{*}{MoViNet-A2} 
     & 64.88  & 62.03 & 60.19 & 62.58 & 62.30  & 62.38 &  63.99& \cellcolor{gray!20}\textbf{62.75} \\
     Noun & & 47.18 & 44.65 & 43.20 & 43.61 & 43.80  & 43.68 & 47.71 & \cellcolor{gray!20}\textbf{46.50} \\

\hline
Verb
&\multirow{2}{*}{SlowFast-R50} 
    & 64.88  & 63.99 & 63.62 & 64.38 & 64.55  & 63.50 & 63.99& \cellcolor{gray!20}\textbf{65.83} \\
     Noun& & 47.18  & 47.71 & 48.76 & 48.48 & 48.96 & 48.87  & 47.71&\cellcolor{gray!20}\textbf{49.80} \\

\hline


\end{tabular}
}
\end{table*}

\section{Experiments}
\subsection{Datasets and Compared Methods}

We evaluate our method on three widely used video action recognition benchmarks:

\textbf{HMDB51}~\cite{kuehne2011hmdb} contains 6,766 video clips spanning 51 action classes (e.g., jump, run, laugh), with at least 101 videos per class. The dataset is split into three standard train/test splits (70/30 per class), and we follow the official splits for training and evaluation.

\textbf{EPIC-KITCHENS-100}~\cite{epic-kitchens} is an egocentric video dataset comprising 100 hours of first-person footage recorded by 45 participants in kitchens. It includes 90K action segments labeled with 97 verbs and 300 nouns. We follow the official split and report Top-1 accuracy on both verb and noun classification.

\textbf{Kinetics-400}~\cite{kay2017kinetics} is a large-scale dataset with ~240K training and 20K validation videos across 400 action classes. Each clip lasts around 10 seconds and is collected from YouTube, offering high diversity. We train on the official training set and evaluate on the validation set.

We focus on three lightweight and efficient CNN architectures as student models: \textbf{MoViNet}~\cite{kondratyuk2021movinets}, \textbf{X3D}~\cite{feichtenhofer2020x3d}, and \textbf{SlowFast}~\cite{ feichtenhofer2019slowfast}. The ViT teacher is \textbf{MViTv2}~\cite{li2022mvitv2}, and the CNN teacher shares the same architecture as the student. The compared state-of-the-art video knowledge distillation methods include: \textbf{ATKD}~\cite{zagoruyko2016paying}, \textbf{MGD}~\cite{yang2022masked}, \textbf{DKD}~\cite{zhao2022decoupled}, and \textbf{PixelKD}~\cite{10579049}, covering diverse paradigms of feature-based distillation and logits-based distillation.

\subsection{Implementation details}
All models are initialized with standard ImageNet and Kinetics-400 pre-trained weights. We train using the SGD optimizer with a batch size of 64, an initial learning rate of 0.01, and a MultiStepLR scheduler. The input clip length is set to 16 frames. Teachers are fixed during distillation. More details are put in the Appendix.

\subsection{Comparisons with State-of-the-arts}

\begin{table}[t]
\centering
\renewcommand{\arraystretch}{1.2}
\caption{
State-of-the-art comparison on Kinetics-400. ViT Teacher: MViTv2-S, CNN Teacher/Student: I3D-R50.
}
\vspace{-5pt}
\label{tab:k400}
\scalebox{0.8}{
\label{tab:kinetics400_comparison}
\setlength{\tabcolsep}{10pt}
\begin{tabular}{lcc}
 \hline
\textbf{Type} & \textbf{Method} & \textbf{Acc (\%)} \\
\hline
\multirow{4}{*}{Only ViT Teacher}
 & ATKD     & 72.06     \\
& MGD      & 73.05  \\
& DKD      & 73.38  \\                           
& PixelKD  & 73.14  \\
\hline
\multirow{1}{*}{ViT Teacher+CNN Teacher}

& \textbf{Ours}    & \textbf{74.16} \\
\hline
\end{tabular}
}
\end{table}

\subsubsection{Results on HMDB51}
Table \ref{tab:hmdb51} reads that our dual-teacher cross-architecture distillation consistently outperforms existing methods across all lightweight CNN students. For SlowFast-R50, it improves accuracy by 5.95\% over the baseline and 3.2\% over the previous best (75.23\% vs. 72.03\%). On MoViNet-A2 and X3D-S, our method achieves 76.41\% and 77.06\%, respectively. We explore two CNN teacher settings: using the same model as the student (e.g., MoViNet-A2, SlowFast-R50) and a larger variant of the same architecture (e.g., X3D-M), both yielding substantial gains. Remarkably, the distilled X3D-S even surpasses its ViT teacher (77.06\% vs. 76.93\%), demonstrating the strong generalization ability of our approach.

\subsubsection{Results on EPIC-KITCHENS-100}
As shown in Table~\ref{tab:epic}, we evaluate our method on the EPIC-KITCHENS-100 dataset, which features egocentric videos with dense, fine-grained action annotations. Our dual-teacher framework consistently outperforms all compared distillation methods across both verb and noun classification tasks. With the MoViNet-A2 backbone, our method achieves 46.50\% noun accuracy, improving over the best single-teacher results by up to 2.70\%. On SlowFast-R50, the student reaches 65.83\% for verbs and 49.80\% for nouns, outperforming all baselines and even the ViT teacher itself (64.88\% and 47.18\%). These results demonstrate that leveraging both ViT and CNN teachers enables more effective knowledge transfer, benefiting from their complementary strengths in modeling global context and architectural alignment.

\subsubsection{Results on Kinetics-400}
Table~\ref{tab:k400} presents the results of various single-teacher distillation methods using MViT-S to distill I3D-R50. Our dual-teacher framework, which combines MViT-S as the ViT teacher and I3D-R50 as the CNN teacher, outperforms all compared feature-, logit-, and hybrid-based approaches. The relatively modest improvement on Kinetics-400 can be attributed to its data characteristics—most videos are short, centered, and less reliant on global context modeling, which aligns well with the strengths of CNNs. Nonetheless, our method still brings consistent improvements across datasets, demonstrating its general applicability and robustness.

\subsection{Ablation Studies}
\label{ssec:ablation}

\subsubsection{Effectiveness of Discrepancy-Aware Teacher Weighting (DATW)}
We perform ablation studies with results summarized in Table~\ref{tab:ablation_dasr}. Specifically, we compare DATW against the following three baseline strategies: \textbf{1) Uniform}, which assigns equal weight to all samples and serves as a static baseline; \textbf{2) Teacher Confidence-only}, which uses only the teacher's confidence score $\mathcal{C}_k(x)$ for weighting; and \textbf{3) Student-Teacher Discrepancy-only}, which uses only the prediction discrepancy between student and teacher $\mathcal{D}_k(x)$. The results show that both \textbf{Confidence-only} and \textbf{Discrepancy-only} clearly outperform the \textbf{Uniform} baseline, indicating that dynamic, sample-specific weighting improves knowledge transfer. Our full DASR mechanism, which integrates both confidence and discrepancy, achieves the best performance of \textbf{73.99\%}. This confirms that combining reliability (confidence) and novelty (discrepancy) yields the most effective supervision during distillation.

\begin{table}[t]
\centering
\caption{Ablation study for our Discrepancy-Aware Teacher Weighting (DATW) mechanism. Our full method, which combines both Teacher Confidence $\mathcal{C}_k(x)$ and Student-Teacher Discrepancy $\mathcal{D}_k(x)$, significantly outperforms other strategies. (\textit{The SDB module was deactivated for this study.})}
\vspace{-7pt}
\scalebox{0.8}{
\setlength{\tabcolsep}{10pt}
\begin{tabular}{lc}
\toprule
\textbf{Method} & \textbf{Accuracy (\%)} \\
\midrule
MViTv2-S (ViT Teacher)       & 76.93 \\
SlowFast-R50 (CNN Teacher)   & 69.28 \\
SlowFast-R50 (CNN Student)   & 69.28 \\
\midrule
Average                     & 71.63 \\
Teacher Confidence: $\mathcal{C}_k(x)$  & 73.86 \\
Stu–Tea Discrepancy: $\mathcal{D}_k(x)$  & 73.20 \\
DATW: $\mathcal{C}_k(x) \cdot \mathcal{D}_k(x)$ & \textbf{73.99} \\
\bottomrule
\end{tabular}
}
\label{tab:ablation_dasr}
\end{table}

\begin{table}[t]
\centering
\caption{Ablation study of the Structure Discrepancy Branch (SDB) on HMDB51. The Non-local Block provides a larger gain due to its global modeling capability.}
\vspace{-7pt}
\label{tab:ablation_sdb}
\scalebox{0.8}{
\setlength{\tabcolsep}{10pt}
\begin{tabular}{lc}
\toprule
\textbf{Method} & \textbf{Accuracy (\%)} \\
\midrule
Baseline (w/o SDB)                  & 75.23 \\
SDB (SENet only)               & 76.14 \\
SDB (Non-Local Block only)     & 76.54 \\
\textbf{SDB (Non-Local + SENet)} & \textbf{77.06} \\
\bottomrule
\end{tabular}
}
\end{table}

\begin{table}[t]
\centering
\caption{Results with different teacher combinations on HMDB51. Greatest gain is achieved when the CNN teacher is structurally homogeneous with the student (X3D-M with X3D-S), highlighting the benefit of architectural alignment.}
\vspace{-7pt}
\label{tab:generalization}
\scalebox{0.8}{
\setlength{\tabcolsep}{12pt}
\begin{tabular}{cccc}
\toprule
ViT Teacher & CNN Teacher & CNN Student & Acc (\%) \\
\midrule
-- & -- & X3D-S & 72.42 \\

\rowcolor{gray!20} 
& R(2+1)D & X3D-S & 75.36 \\
\rowcolor{gray!20} 
\multirow{-2}{*}{Swin-B} & X3D-M & X3D-S & 76.80 \\

\multirow{2}{*}{MViTv2-S} 
& R(2+1)D & X3D-S & 74.77 \\
& X3D-M & X3D-S & \textbf{77.06} \\
\bottomrule
\end{tabular}
}
\end{table}

\begin{table}[t]
\centering
\caption{Comparison of model complexity and accuracy on HMDB51. Our method establishes a superior performance-efficiency trade-off, enabling lightweight students to achieve high accuracy while maintaining computational efficiency.}
\vspace{-7pt}
\resizebox{\linewidth}{!}{
\begin{tabular}{lllll}
\toprule
Arch. & Model           & FLOPs (G)  & Params (M) & Acc (\%) \\
\midrule
\multirow{4}{*}{ViT} 
& MViTv2-S        & 64.45     & 34.50      & 76.93       \\
& UniFormerv2-B   & 96.74     & 49.81      & 73.59       \\
& TimeSformer     & 180       & 86.11      & 72.81       \\
& Swin-B          & 282       & 88.05      & 75.42       \\
\midrule
\multirow{8}{*}{CNN} 
& I3D-R50         & 33.27     & 27.33      & 71.05       \\
& R(2+1)D         & 213       & 63.58      & 74.12       \\
& \cellcolor{gray!20}SlowFast-R50    & \cellcolor{gray!20}36.10     & \cellcolor{gray!20}34.40      & \cellcolor{gray!20}69.28       \\
& \cellcolor{gray!20}\textbf{SlowFast-R50 (Ours)} & \cellcolor{gray!20}36.10 & \cellcolor{gray!20}34.40 &\cellcolor{gray!20}\textbf{75.23} \\
& MoViNet-A2      & 3.85      & 4.80       & 71.76       \\
& \textbf{MoViNet-A2 (Ours)}   & 3.85  & 4.80  & \textbf{76.41} \\ 
& \cellcolor{gray!20}X3D-S           & \cellcolor{gray!20}2.45      & \cellcolor{gray!20}3.76       & \cellcolor{gray!20}72.42       \\
& \cellcolor{gray!20}\textbf{X3D-S (Ours)}        &\cellcolor{gray!20}2.45  &\cellcolor{gray!20}3.76  & \cellcolor{gray!20}\textbf{77.06} \\
\bottomrule
\end{tabular}
}
\label{tab:FLOPs_Params}
\end{table}

\subsubsection{Effectiveness of the Structure Discrepancy Branch (SDB)}
To evaluate the effectiveness of different instantiations of SDB, we compare three variants: \textbf{1) SDB with SE module}, \textbf{2) SDB with Non-local block}, and \textbf{3) SDB with both}. As shown in Table~\ref{tab:ablation_sdb}, no matter how the SDB is instantiated—whether using a Non-local block or an SE module—its inclusion consistently improves student performance over the baseline (75.23\%), confirming the value of leveraging the residual between ViT and CNN features as a supervision signal. Among the instantiations, using a Non-local block brings the accuracy to 76.54\% (+1.31\%), while the SE module achieves 76.14\% (+0.91\%). This shows that while both are effective, the Non-local block provides stronger gains due to its ability to model long-range dependencies, which aligns well with the global context discrepancy captured by ViT-CNN residuals. Combining both leads to the highest performance of 77.06\% (+1.83\%), suggesting that SE and Non-local modules contribute complementary inductive biases. These results collectively validate both the design of SDB and our choice to model ViT-specific global structural cues as a learning target.

\subsubsection{Generalization Study with Different Teachers.}
To assess the generalization capability of our dual-teacher distillation framework, we conduct a study using four different teacher combinations on the HMDB51 dataset. In all settings, we fix the student model as X3D-S to ensure fair comparison and isolate the impact of teacher pairings. Specifically, we experiment with two ViT teachers (Swin-B and MViTv2-S) and two CNN teachers (R(2+1)D and X3D-M), covering both heterogeneous and homogeneous architectures. As shown in Table~\ref{tab:generalization}, our framework consistently improves the student’s performance across all teacher combinations. The highest accuracy of \textbf{77.06\%} is achieved when the CNN teacher (X3D-M) is structurally aligned with the student (X3D-S), underscoring the importance of architectural consistency in effective knowledge transfer. 

\subsubsection{Comparison of Model Complexity and Performance.}
Table~\ref{tab:FLOPs_Params} shows that, with our framework, X3D-S improves by 4.64\% to \textbf{77.06\%} without any inference overhead—surpassing not only other lightweight models but also its ViT teacher (i.e., MViTv2-S). MoViNet-A2 and SlowFast-R50 show similar gains. These results confirm that our approach can effectively overcome the performance bottleneck of lightweight CNNs, enabling them to rival or exceed larger ViT models at lower cost.

\section{Conclusion}
This work has tackled two core limitations of cross-architecture knowledge distillation (CAKD) for video action recognition: the over-reliance on ViT supervision and the neglect of structurally aligned CNN teachers. We have proposed a dual-teacher framework that integrates the complementary strengths of both architectures through adaptive teacher weighting and structure discrepancy-aware distillation. These strategies jointly address the key challenges of supervision balancing and heterogeneous knowledge transfer. Extensive experiments show that our method consistently enhances lightweight CNN students, outperforming both teachers and existing distillation approaches. While we have demonstrated strong results with fixed teachers, future work could explore joint optimization or dynamic teacher adaptation to further improve flexibility and effectiveness.


\bibliography{aaai2026}

\end{document}